\documentclass[10pt,twocolumn,letterpaper]{article}

\pdfoutput=1

\usepackage{booktabs}
\usepackage{pifont}
\usepackage{cvpr}
\usepackage{times}
\usepackage{epsfig}
\usepackage{graphicx}
\usepackage{amsmath}
\usepackage{amssymb}
\usepackage{pbox}
\usepackage{siunitx}
\usepackage{gensymb}
\usepackage{verbatim}
\usepackage{verbatimbox}
\usepackage{fancyvrb}

\usepackage[table]{xcolor}
\definecolor{bedColor}{rgb}{0, 0, 1}
\definecolor{booksColor}{rgb}{0.9137,0.3490,0.1882}
\definecolor{ceilColor}{rgb}{0, 0.8549, 0}
\definecolor{chairColor}{rgb}{0.5843,0,0.9412}
\definecolor{floorColor}{rgb}{0.8706,0.9451,0.0941}
\definecolor{furnColor}{rgb}{1.0000,0.8078,0.8078}
\definecolor{objsColor}{rgb}{0,0.8784,0.8980}
\definecolor{paintColor}{rgb}{0.4157,0.5333,0.8000}
\definecolor{sofaColor}{rgb}{0.4588,0.1137,0.1608}
\definecolor{tableColor}{rgb}{0.9412,0.1373,0.9216}
\definecolor{tvColor}{rgb}{0,0.6549,0.6118}
\definecolor{wallColor}{rgb}{0.9765,0.5451,0}
\definecolor{windColor}{rgb}{0.8824,0.8980,0.7608}

\newcommand{\xmark}{\ding{55}}
\newcommand{\cmark}{\ding{51}}%

\usepackage{listings}
\definecolor{codegreen}{rgb}{0,0.6,0}
\definecolor{codegray}{rgb}{0.5,0.5,0.5}
\definecolor{codepurple}{rgb}{0.58,0,0.82}
\definecolor{backcolour}{rgb}{0.95,0.95,0.92}

\lstdefinestyle{mystyle}{
    backgroundcolor=\color{backcolour},   
    commentstyle=\color{codegreen},
    keywordstyle=\color{magenta},
    numberstyle=\tiny\color{codegray},
    stringstyle=\color{codepurple},
    basicstyle=\scriptsize\ttfamily,
    breakatwhitespace=false,         
    breaklines=true,                 
    captionpos=b,                    
    keepspaces=true,                 
    numbers=left,                    
    numbersep=5pt,                  
    showspaces=false,                
    showstringspaces=false,
    showtabs=false,                  
    tabsize=2
}

\usepackage[breaklinks=true,bookmarks=false]{hyperref}

\hypersetup{
  colorlinks   = true, 
  urlcolor     = red, 
  linkcolor    = red, 
  citecolor   =  green 
}

\cvprfinalcopy 

\def\cvprPaperID{****} 

\begin{document}

\title{SceneNet RGB-D: 5M Photorealistic Images of\\ Synthetic Indoor Trajectories with Ground Truth}

\author{John McCormac,
Ankur Handa,
Stefan Leutenegger,
Andrew J. Davison\\
Dyson Robotics Laboratory at Imperial College, Department of Computing,\\
Imperial College London \\
{\tt\small \{brendan.mccormac13,s.leutenegger,a.davison\}@imperial.ac.uk, handa.ankur@gmail.com}
}

\twocolumn[{%
\renewcommand\twocolumn[1][]{#1}%
\maketitle
\centerline{
{
\includegraphics[width=0.24\linewidth]{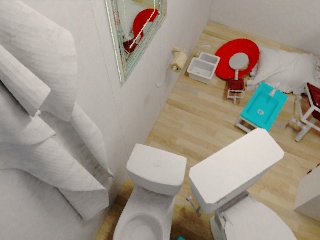}
\hspace{-0.6em}
\includegraphics[width=0.24\linewidth]{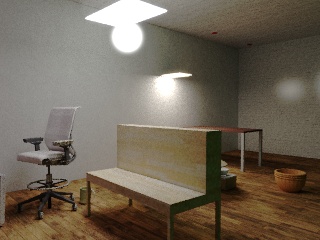}
\hspace{-0.6em}
\includegraphics[width=0.24\linewidth]{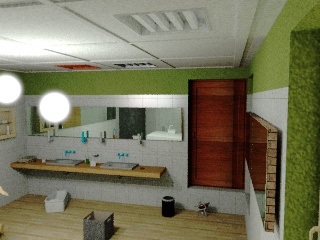}
\hspace{-0.6em}
\includegraphics[width=0.24\linewidth]{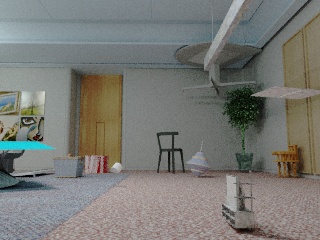}
\hspace{-0.6em}
}
}
\vspace{-0.5em}
\centerline{
{
\includegraphics[width=0.24\linewidth]{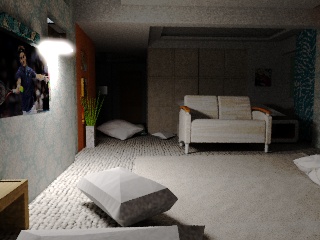}
\hspace{-0.6em}
\includegraphics[width=0.24\linewidth]{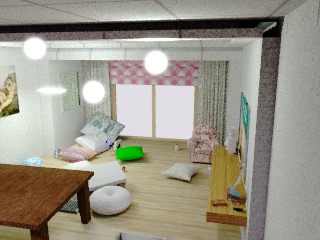}
\hspace{-0.6em}
\includegraphics[width=0.24\linewidth]{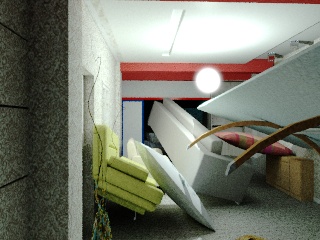}
\hspace{-0.6em}
\includegraphics[width=0.24\linewidth]{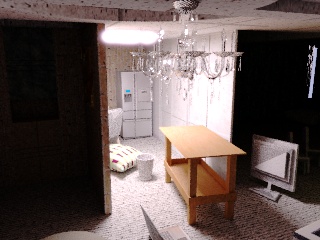}
\hspace{-0.6em}
}
}

\hfill
\hfill
}
]

\begin{abstract}
 
\vspace{-1.0em}
We introduce SceneNet RGB-D, expanding the previous work of SceneNet to enable large scale photorealistic rendering of indoor scene trajectories. It provides pixel-perfect ground truth for scene understanding problems such as semantic segmentation, instance segmentation, and object detection, and also for geometric computer vision problems such as optical flow, depth estimation, camera pose estimation, and 3D reconstruction. Random sampling permits virtually unlimited scene configurations, and here we provide a set of 5M rendered RGB-D images from over 15K trajectories in synthetic layouts with random but physically simulated object poses. Each layout also has  random lighting, camera trajectories, and textures. The scale of this dataset is well suited for pre-training data-driven computer vision techniques from scratch with RGB-D inputs, which previously has been limited by relatively small labelled datasets in NYUv2 and SUN RGB-D.  It also provides a basis for investigating 3D scene labelling tasks by providing perfect camera poses and depth data as proxy for a SLAM system. We host the dataset at \url{http://robotvault.bitbucket.io/scenenet-rgbd.html}.

\end{abstract}

\section{Introduction}

\begin{table*}[t] 
\centering
\resizebox{2.1\columnwidth}{!}{%
\begin{tabular}{l l l l l l l l} \toprule
  & Stanford Scenes & NYUv2 & SUN RGB-D & \hspace{7mm} SceneNet & sceneNN & \hspace{5mm} SUN CG$^{*}$ & \textbf{SceneNet RGB-D}   \\ \midrule
RGB-D videos available & \multicolumn{1}{c}{\xmark} & \multicolumn{1}{c}{\cmark} & \multicolumn{1}{c}{\xmark} & \multicolumn{1}{c}{\xmark} & \multicolumn{1}{c}{\cmark} & \multicolumn{1}{c}{\xmark} &  \multicolumn{1}{c}{\cmark}\\ 
Per-pixel annotations & \multicolumn{1}{c}{NA} & \multicolumn{1}{c}{Key frames} & \multicolumn{1}{c}{Key frames} & \multicolumn{1}{c}{Key frames} & \multicolumn{1}{c}{Videos} & \multicolumn{1}{c}{Key Frames} &  \multicolumn{1}{c}{Videos}\\ 
Trajectory ground truth & \multicolumn{1}{c}{\xmark} & \multicolumn{1}{c}{\xmark} & \multicolumn{1}{c}{\xmark} & \multicolumn{1}{c}{\xmark} & \multicolumn{1}{c}{\cmark} & \multicolumn{1}{c}{\xmark} &  \multicolumn{1}{c}{\cmark}\\
RGB texturing & \multicolumn{1}{c}{Non-photorealistic} & \multicolumn{1}{c}{Real} & \multicolumn{1}{c}{Real} & \multicolumn{1}{c}{Non-photorealistic} & \multicolumn{1}{c}{Real} & \multicolumn{1}{c}{Non-photorealistic} & \multicolumn{1}{c}{Photorealistic}\\ 
Number of layouts & \multicolumn{1}{c}{1723} & \multicolumn{1}{c}{464} & \multicolumn{1}{c}{-} & \multicolumn{1}{c}{57} & \multicolumn{1}{c}{100} & \multicolumn{1}{c}{45,622} & \multicolumn{1}{c}{57} \\
Number of configurations & \multicolumn{1}{c}{1723} & \multicolumn{1}{c}{464} & \multicolumn{1}{c}{-} & \multicolumn{1}{c}{1000} & \multicolumn{1}{c}{100} & \multicolumn{1}{c}{45,622} & \multicolumn{1}{c}{16,895} \\
3D models available & \multicolumn{1}{c}{\cmark} & \multicolumn{1}{c}{\xmark} & \multicolumn{1}{c}{\xmark} & \multicolumn{1}{c}{\cmark} & \multicolumn{1}{c}{\cmark} & \multicolumn{1}{c}{\cmark} & \multicolumn{1}{c}{\cmark} \\ 
Method of design & \multicolumn{1}{c}{Manual} & \multicolumn{1}{c}{Real} & \multicolumn{1}{c}{Real} & \multicolumn{1}{c}{Manual and Random} & \multicolumn{1}{c}{Real} & \multicolumn{1}{c}{Manual} & \multicolumn{1}{c}{Random} \\ \bottomrule
 \hline
\end{tabular}
}
\caption{A comparison table of indoor scene datasets and their differing characteristics. SceneNN is an example of a real world dataset that does provide 3D models, however the models are not water-tight. Stanford Scenes database \cite{Fisher:etal:SIGGRAPHASIA2012} does not provide any explicit ground truth as it is primarily designed for scene retrieval. SUN RGB-D captures short video clips but only release single image annotations. *At the time of this manuscript's publication the SUN CG dataset was not released. They provide dense volumetric annotations, however only for single images.}
\label{table:different datasets}
\end{table*}

A primary goal of computer vision research is to give computers the capability to reason about real world images in a human-like manner.  This includes a semantic understanding of the objects present in the scene, their locations, 6-DoF poses, and an intuitive grasp of the physics involved. Recent years have witnessed a huge interest in scene understanding, largely sparked by the seminal work of AlexNet \cite{Krizhevsky:etal:NIPS2012} and the increasing popularity of Convolutional Neural Networks (CNNs). That work highlighted the importance of large scale labelled datasets when working with data-hungry supervised learning algorithms. In this work we focus on the challenge of obtaining large quantities of labelled data with the aim of alleviating the need for collecting datasets through manual effort. 

In particular, we are motivated by tasks which require more than a simple text label for an image.  For tasks such as semantic labelling and instance segmentation, obtaining accurate per-pixel ground truth annotations by hand is prohibitively expensive. In other cases it can be almost impossible, such as for fine-grained optical flow data, or metrically accurate 3D pose information for an object.  Inspired by the recent success of synthetic data for training scene understanding systems, our goal has been to generate a large scale dataset of photorealistic RGB-D videos which provide perfect and complete ground truth for a wide range of problems.

Our dataset has several key strengths relative to other publicly
available datasets for indoor scene understanding that
make it especially useful for training computer vision models, which could be used for real-world applications in robotics and augmented reality. We have used ray-tracing to generate
high quality synthetic RGB images, aiming towards photorealism with
full lighting effects and elements such as motion blur, as well as accompanying synthetic depth images.
The images are rendered from
randomly generated smooth trajectories to create sequential video
clips from a moving virtual camera, opening up research on temporal
fusion for high quality labelling. Our process to generate the
contents of the synthetic scenes observed has relied to the greatest
degree possible on fully automatic methods, with object distributions statistically sampled from publicly available real-world scene repositories and randomly positioned within a physics simulation that then ensures feasible configurations. This means that our pipeline can produce a greater degree of variety of scene configurations than others, enabling a potentially much larger dataset without the need for human scene design or annotation.

In Section~\ref{sec:dataset_overview} we discuss the overall dataset pipeline and available ground truth labels. In Section~\ref{sec:metric scales} below, we describe the process of obtaining metric scales of objects from SUN RGB-D. Section~\ref{sec:generating_random_scenes_with_physics_engine} provides a detailed explanation on random scene generation, and Section~\ref{ref:random trajectories} talks about generating random trajectories.

\begin{figure*}[!htb]
\centerline{
\hfill { \includegraphics[width=1.01\linewidth]{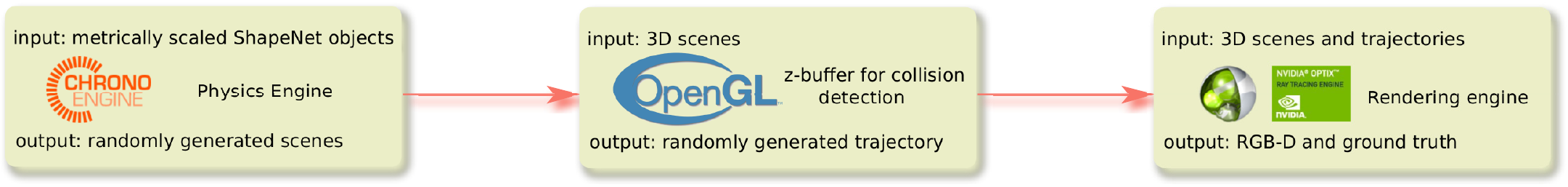} }
}
\caption{Flow chart of the different stages in our pipeline. Physically realistic scenes are created using Chrono Engine by dropping objects from the ceiling. These scenes are used to generate automated camera trajectories simulating human hand-held motion and both are passed on the rendering engine --- inspired by OptiX --- to produce {RGB-D} ground truth.}
\label{fig:flow_chart}
\end{figure*}

\section{Background}

A growing body of research has highlighted that carefully synthesised artificial data with appropriate noise models can be an effective substitute for real-world labelled data in areas that ground-truth data is difficult to obtain. Aubry \textit{et al.} \cite{Aubry:etal:CVPR2014} used synthetic 3D CAD models for learning visual elements to do 2D-3D alignment in images, and similarly, Gupta \textit{et al.} \cite{Gupta:etal:CVPR2015} trained on rendering of synthetic objects to do alignment of 3D models with RGB-D images. Peng~\textit{et al.}\cite{Peng:etal:arXiv2014} augmented small datasets of objects with renderings of synthetic 3D objects with random textures and backgrounds to improve object detection performance. FlowNet \cite{Fischer:etal:arXiv2015} and recently FlowNet 2.0 \cite{Ilg:etal:arXiv2016} showed that remarkable improvements can be made with training data obtained from synthetic scenes for optical flow estimation. de Souza \textit{et al.} \cite{deSouza:etal:arXiv2016} use procedural generation of human actions with computer graphics to generate large dataset of videos for human action recognition.

As a precursor to the present work, Handa \textit{et al.} \cite{Handa:etal:ARXIV2015} produced SceneNet, a repository of labelled synthetic 3D scenes from five different categories.  This repository was used to generate per-pixel semantic segmentation ground truth for depth only images from random viewpoints. They demonstrate that a network trained on 10K images of synthetic depth data and fine-tuned on the original NYUv2 \cite{Silberman:etal:ECCV2012} and SUN RGB-D \cite{Song:etal:CVPR2015} datasets shows an increase in the performance on the task of semantic segmentation when compared to the network trained on just the original datasets. 

For outdoor scenes, Ros~\textit{et al.} generated the SYNTHIA \cite{Ros:etal:CVPR16} dataset for road scene understanding, and work by Richter \textit{et al.} \cite{Richter:etal:ECCV2016} produced synthetic training data from a photorealistic gaming engine.  This is an exciting avenue, however it is not always possible to obtain the required data from gaming engines, which due to proprietary issues lack the flexibility of a fully open-source alternative. SceneNet {RGB-D} uses open-source scene layouts \cite{Handa:etal:ARXIV2015} and 3D object repositories \cite{Shapenet:etal:2015} that provide textured objects.  We have also built upon an open-source ray-tracing framework which allows significant flexibility in the ground truth data we can collect and visual effects we can simulate.

For indoor scenes, recent work by Qui~\textit{et al.} \cite{Qiu:etal:2016} called UnrealCV provides a plugin to generate ground truth data and photorealistic images from UnrealEngine. However, they do not provide any labelled dataset and their plugin uses scene assets created by artists, which assists in the apparent photorealism with high quality assets.  Assets of this quality are often proprietary, and difficult to source in large quantities.  Finally, they do not explore random scene generation systems as we do here. 

Our dataset, SceneNet {RGB-D}, samples random layouts from SceneNet \cite{Handa:etal:ARXIV2015} and objects from ShapeNet \cite{Shapenet:etal:2015} to create a potentially unlimited number of scene configurations. As shown in Table~\ref{table:different datasets}, there are a number of key differences between our work and other available datasets. It is one of the first to provide large quantities of photorealistic renderings of indoor scenes.  Hua~\textit{et al.} provide sceneNN \cite{Hua:etal:3DV2016}, a dataset of 100 labelled meshes of real world scenes, obtained with a reconstruction system with objects labelled directly in 3D for semantic segmentation ground truth. Such real-world datasets are often limited in scale due to the amount of manual effort required.

\begin{figure*}[!htb]
\vspace{2mm}
\centerline{
\hfill { \includegraphics[width=0.19\linewidth]{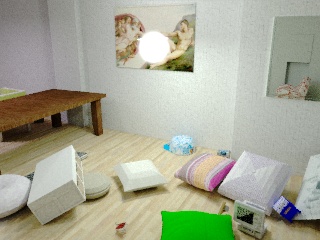} }
\hfill { \includegraphics[width=0.19\linewidth]{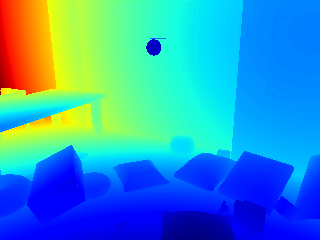} }
\hfill { \includegraphics[width=0.19\linewidth]{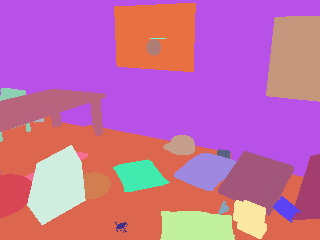} }
\hfill { \includegraphics[width=0.19\linewidth]{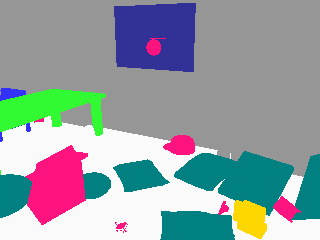} }
\hfill { \includegraphics[width=0.19\linewidth]{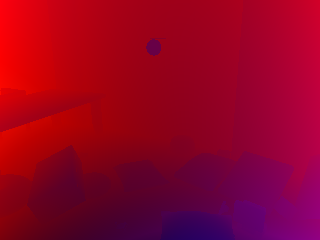} }
\hfill
}
\vspace{2mm}
\centerline{
\hfill { \includegraphics[width=0.19\linewidth]{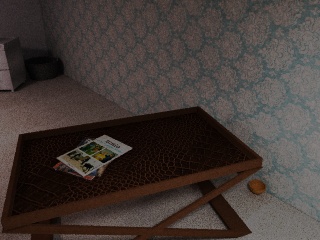} }
\hfill { \includegraphics[width=0.19\linewidth]{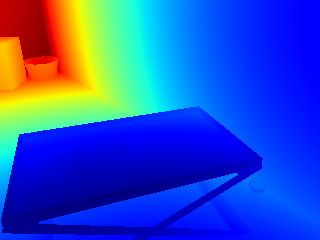} }
\hfill { \includegraphics[width=0.19\linewidth]{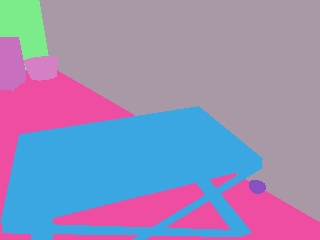} }
\hfill { \includegraphics[width=0.19\linewidth]{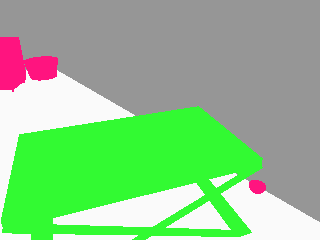} }
\hfill { \includegraphics[width=0.19\linewidth]{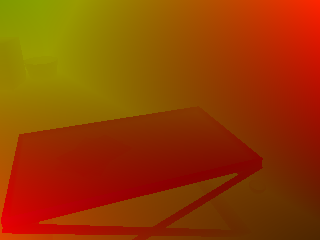} }
\hfill
}
\vspace{2mm}
\centerline{
\hfill { \includegraphics[width=0.19\linewidth]{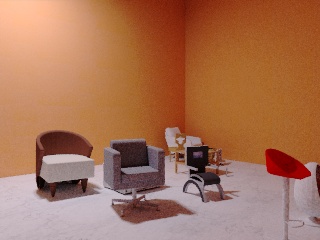} }
\hfill { \includegraphics[width=0.19\linewidth]{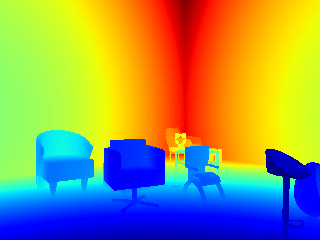} }
\hfill { \includegraphics[width=0.19\linewidth]{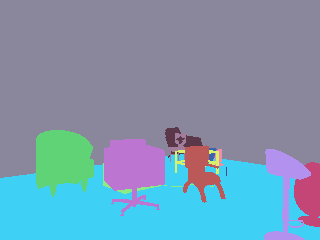} }
\hfill { \includegraphics[width=0.19\linewidth]{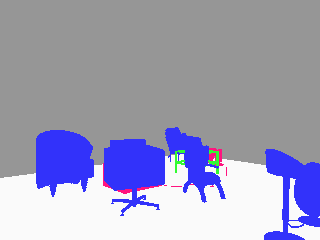} }
\hfill { \includegraphics[width=0.19\linewidth]{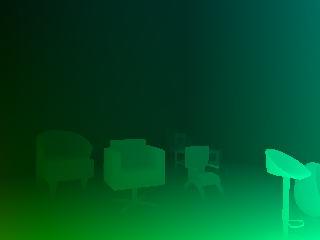} }
\hfill
}
\vspace{2mm}
\centerline{
\hfill { \includegraphics[width=0.19\linewidth]{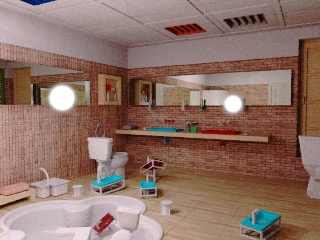} }
\hfill { \includegraphics[width=0.19\linewidth]{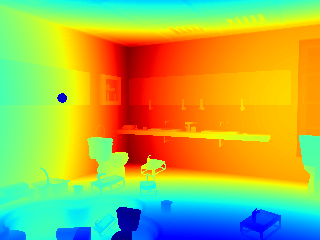} }
\hfill { \includegraphics[width=0.19\linewidth]{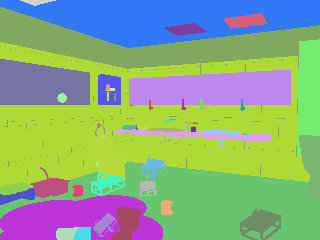} }
\hfill { \includegraphics[width=0.19\linewidth]{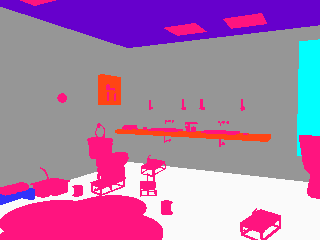} }
\hfill { \includegraphics[width=0.19\linewidth]{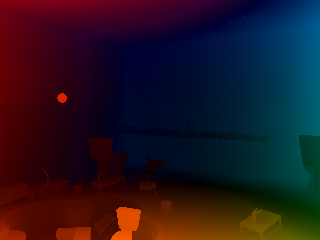} }
\hfill
}
\vspace{2mm}
\centerline{
\hfill { \includegraphics[width=0.19\linewidth]{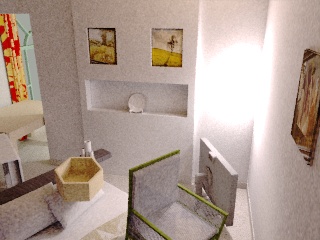} }
\hfill { \includegraphics[width=0.19\linewidth]{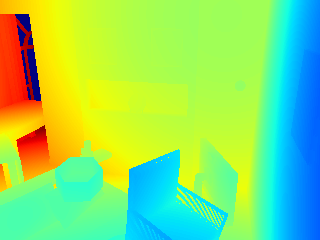} }
\hfill { \includegraphics[width=0.19\linewidth]{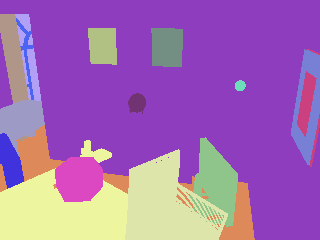} }
\hfill { \includegraphics[width=0.19\linewidth]{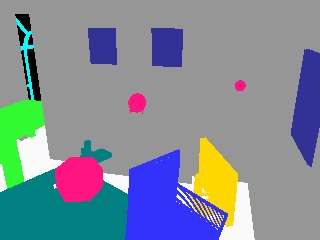} }
\hfill { \includegraphics[width=0.19\linewidth]{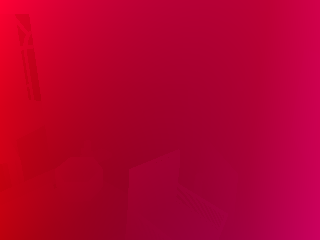} }
\hfill
}
\centerline{
\hfill \makebox[0.19\linewidth][c] {\sf \footnotesize (a) photo}
\hfill \makebox[0.19\linewidth][c] {\sf \footnotesize (b) depth}
\hfill \makebox[0.19\linewidth][c] {\sf \footnotesize (c) instance}
\hfill \makebox[0.19\linewidth][c] {\sf \footnotesize (d) class segmentation}
\hfill \makebox[0.19\linewidth][c] {\sf \footnotesize (e) optical flow}
\hfill
}
\caption{Hand-picked examples from our dataset.  Rendered images on the left and the available ground truth information on the right.}
\label{fig:groundtruth}
\end{figure*}

Recently, Song~\textit{et al.} released the SUN-CG dataset \cite{Song:etal:ARXIV2016} which consists of 45,622 synthetic scene layouts created using Planner5D. There are a few key differences between this and our work.  First, they have not aimed towards rendering photorealistic images of their scenes. Second, our dataset explicitly provides a sequential video trajectory within a scene, allowing 3D correspondences between viewpoints for 3D scene understanding tasks, with the ground truth camera poses acting in lieu of a SLAM system\cite{McCormack:etal:arXiv2016}. Third, their approach to scene generation is quite different. While they have many examples of natural looking manually designed scenes, our approach produces more chaotic configurations that can be generated on-the-fly with almost no chance of repeating. Moreover, since layout textures, positions of light sources, and camera trajectories are all randomised we are able to generate a wide variety of geometrically identical but visually differing renders as shown in Figure \ref{fig:same_scene_diff_settings}. 

We believe such randomness could help prevent overfitting by providing a significantly less predictable set of training examples with high instructional value. It remains an open question whether randomness is preferrable to well designed scenes for learning algorithms, but the recent works of FlowNet and FlowNet 2.0~\cite{Fischer:etal:arXiv2015,Ilg:etal:arXiv2016} seem to suggest that randomness is potentially helpful. Randomness also leads to a simpler data generation pipeline and, given a sufficient computational budget, allows for dynamic on-the-fly generated training examples suitable for active machine learning.  A combination of these two approaches, with a reasonable manually designed scene layouts and added physically simulated noise and clutter may in the end provide the best of both worlds.

\section{Dataset Overview}
\label{sec:dataset_overview}

The overall process from sampling objects to rendering {RGB-D} frames is shown in Figure~\ref{fig:flow_chart}.  For the dataset, we had to balance the competing requirements of frame-rates for video sequences with the computational cost of rendering many very similar images, which would not provide significant variation in the training set for CNNs. We decided upon 5 minute trajectories at 320$\times$240 image resolution, but with a single frame per second, resulting in 300 images per trajectory. Each render takes 2--3 seconds on an NVIDIA GTX 1080 GPU. There is also a trade off between rendering time and quality of renders (See Figure~\ref{fig:renderingtimequality} in Section~\ref{sec:rendering_quality}).

Our trajectory is calculated with a frame-rate of 25Hz, however we only render every 25th pose. Each pose consists of a pair of poses, which define the shutter open and shutter close of the camera. We sample from poses linearly interpolated between the two points to produce motion blur artefacts for simulating any rapid camera shaking. Different ground truth labels can be obtained with an extra rendering pass \textit{e.g.} instance labels are obtained by assigning indices to each object and rendering for each pixel the index of each object instead of RGB values. Depth is defined as the ray length from the camera origin to the first surface it intersects with, this provides perfect depth information even in the case of reflections and motion blur. For ground truth we do not sample multiple points for each pixel as we do for RGB, instead a single ray is emitted from the center of the pixel. 

From these ground truth images it is possible to calculate a number of other pieces of ground truth information.  For example, in accompanying datafiles for each trajectory we store a mapping from each instance label to a semantic label.  These semantic labels are defined with a WordNet id, which provides a useful network structure for semantic links and hierarchies.  In total we have 255 different WordNet semantic categories, including 40 WordNet ids outside of the normal corpus, which were added by the ShapeNet dataset. Given the static scene assumption, the instantaneous optical flow can be calculated using the camera pose along with the depth map. Some examples of the available ground-truth information for a corresponding image is shown in Figure~\ref{fig:groundtruth}.

\begin{figure}[!htb]
\centerline{
\hfill { \includegraphics[width=0.3\linewidth]{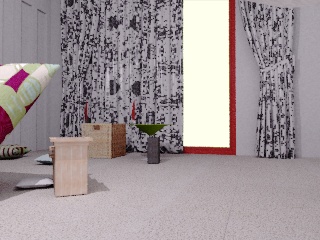} }
\hfill { \includegraphics[width=0.3\linewidth]{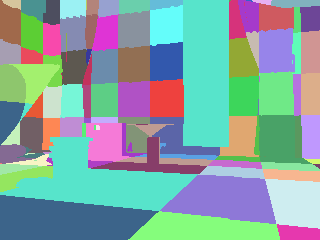} }
\hfill { \includegraphics[width=0.3\linewidth]{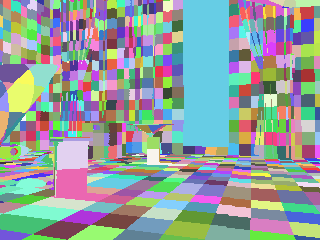} }
}
\centerline{
\hfill \makebox[0.30\linewidth][c] {\sf \footnotesize (a) Photo}
\hfill \makebox[0.30\linewidth][c] {\sf \footnotesize (b) 0.5m voxels}
\hfill
\makebox[0.30\linewidth][c] {\sf \footnotesize (c) 0.15m voxels}
}
\caption{On the left is the original photo, on the right are unique randomly coloured voxels  that remain the same throughout a trajectory. Outside the window there is no depth reading so we assign all of these areas the same default identifier.}
\label{fig:voxel_ground_truth}
\end{figure}

Using the inverse camera model reprojection and the perfect depth map, it is also possible to calculate the 3D position of each surface in the scene.  We use this to calculate voxel correspondence indices (for some arbitrarily selected voxel size) for an entire trajectory, to mimic the correspondences available in a perfect SLAM system.  For an example colorisation of this correspondence system see Figure~\ref{fig:voxel_ground_truth}.

Our dataset is separated into train, validation, and test sets.  Each of these sets has a unique set of layouts, objects, and trajectories particular to the set.  However the parameters for randomly choosing lighting and trajectories remains the same.  We selected two layouts from each type (bathroom, kitchen, office, living room, and bedroom) for the validation and test sets making the layout split 37-10-10. For ShapeNet objects within a scene we randomly divide the objects within each wordnet class into 80-10-10\% splits for train-val-test. This ensures that some of each type of object are in each training set. Our final training set has 5M images from 16K images, our validation and test set have 300K images from 1K different room layouts.  Each layout has a single accompanying trajectory through it. 

\section{Obtaining metric scales of CAD models}
\label{sec:metric scales}

\begin{figure*}[!htb]
\centerline{
\hfill { \includegraphics[width=0.32\linewidth]{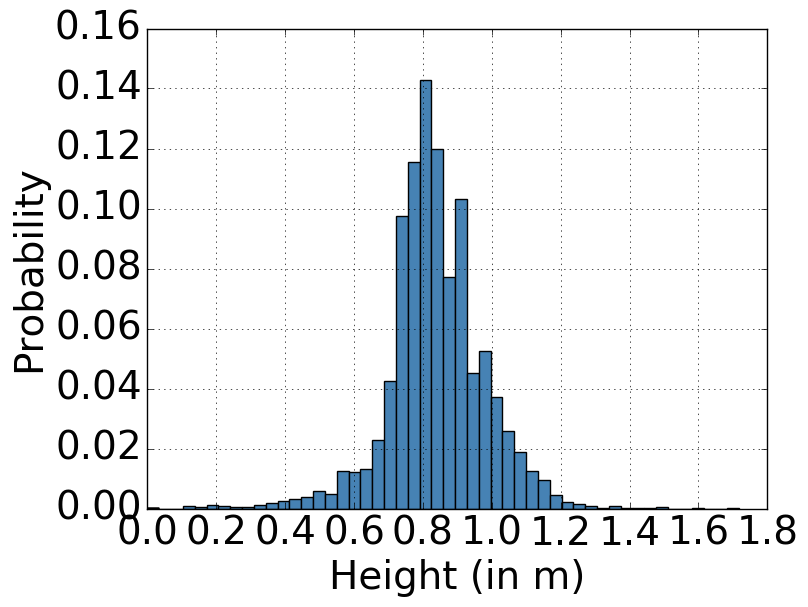} }
\hfill { \includegraphics[width=0.32\linewidth]{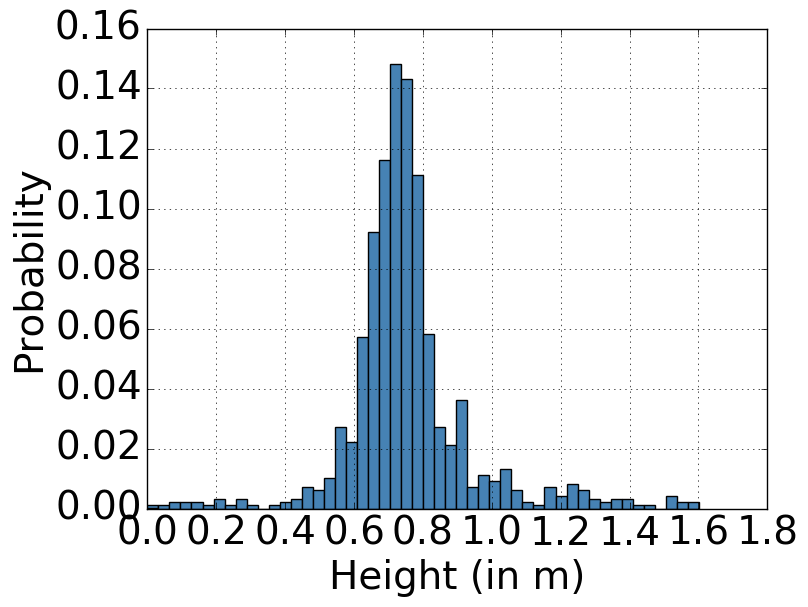} }
\hfill { \includegraphics[width=0.32\linewidth]{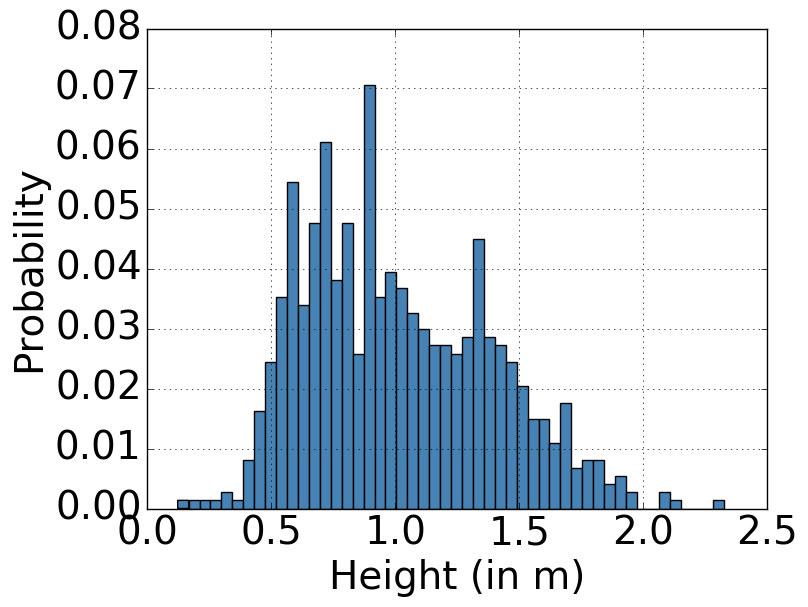} }
\hfill
}
\centerline{
\hfill \makebox[0.30\linewidth][c] {\sf \footnotesize (a) chair}
\hfill \makebox[0.32\linewidth][c] {\sf \footnotesize (b) desk}
\hfill \makebox[0.32\linewidth][c] {\sf \footnotesize (c) bed}
}
\centerline{
\hfill { \includegraphics[width=0.32\linewidth]{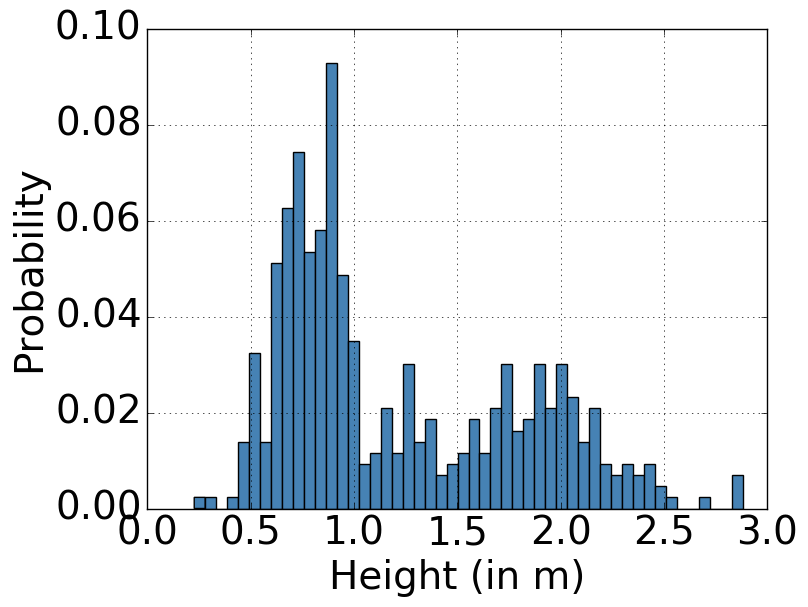} }
\hfill { \includegraphics[width=0.32\linewidth]{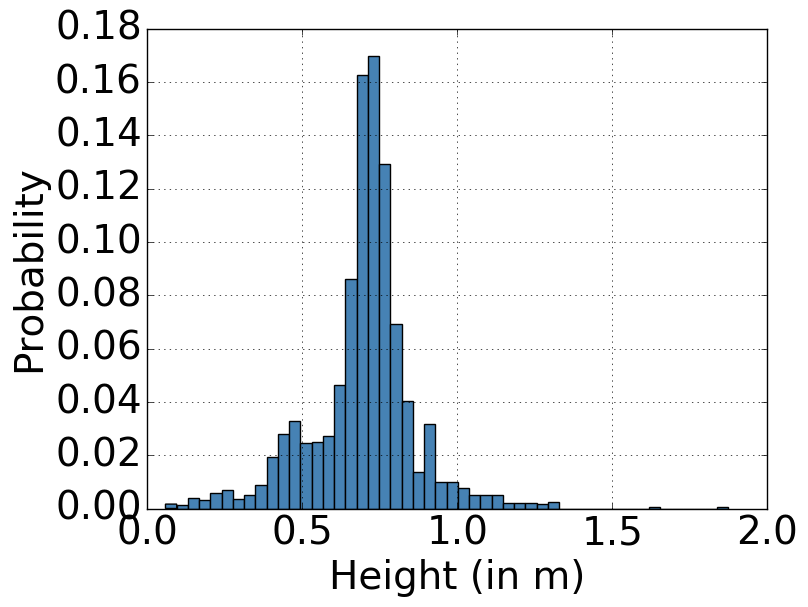} }
\hfill { \includegraphics[width=0.32\linewidth]{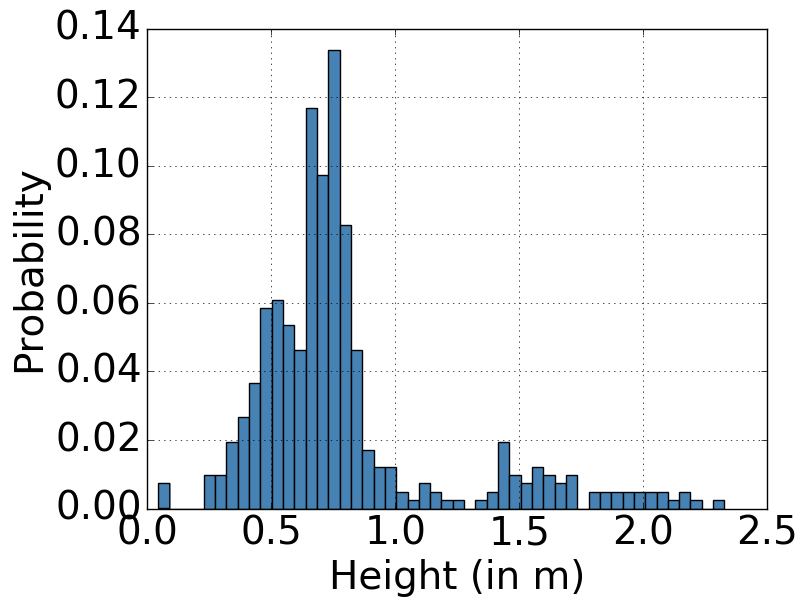} }
\hfill
}
\centerline{
\hfill \makebox[0.30\linewidth][c] {\sf \footnotesize (c) cabinet}
\hfill \makebox[0.32\linewidth][c] {\sf \footnotesize (d) table}
\hfill \makebox[0.32\linewidth][c] {\sf \footnotesize (e) lamp}
}
\caption{Probability distributions of heights (in m) of different objects as obtained from SUN RGB-D. It is interesting to see that some objects like cabinets and lamps clearly do have multimodal height distributions.}
\label{fig:metric_scales}
\end{figure*}

The majority of 3D models in CAD repositories or open-source libraries are created by 3D designers and artists without any explicit designation of metric-scale information. However, it is desirable that the objects placed in our synthetic scenes have similar statistics to their corresponding real world counterparts in terms of their physical dimensions. Fortunately, datasets like SUN RGB-D \cite{Song:etal:CVPR2015} are captured with a depth camera and provide metric 3D bounding boxes for each labelled object in the scene. We leverage this information to obtain the height distribution of object categories, and then randomly sample metric heights from this distribution to scale each object before placing it in the scene. We maintain the aspect ratio of these objects during this scaling procedure. Figure \ref{fig:metric_scales} shows probability distribution of heights of some objects as obtained from SUN RGB-D.

This simple approach is not entirely without drawbacks. The lack of granularity within classes can lead to multimodal height distributions.  For example bedside lamps and floor lamps both are within the same `lamp' class for our purposes, however their heights vary significantly. If the height of a floor lamp is applied to a squat bedside lamp, the resulting object can appear closer in its dimensions to a refrigerator. Tackling this is a significant problem and some work has been done which could be useful in future iterations \cite{Savva:etal:2014}.

\section{Generating random scenes with physics}
\label{sec:generating_random_scenes_with_physics_engine}

We use an off-the-shelf physics engine, Project Chrono\footnote{\url{https://projectchrono.org/}}, to dynamically simulate the scene. We opted for this rather than a computationally more efficient static geometric analysis system for a number of reasons.  Firstly, the computational bottleneck in our system was the rendering pipeline, the physics engine uses the CPU which leaves the GPU free for rendering and can simulate many scenes in the time it takes to render one. Secondly, off-the-shelf physics software was readily available and quite easy to use, and resulted in reasonable looking layouts.  Finally, a full physics simulator leaves open the potential for physically simulated dynamic scenes in future work.

\begin{figure*}[!htb]
\centerline{
\includegraphics[width=0.97\linewidth]{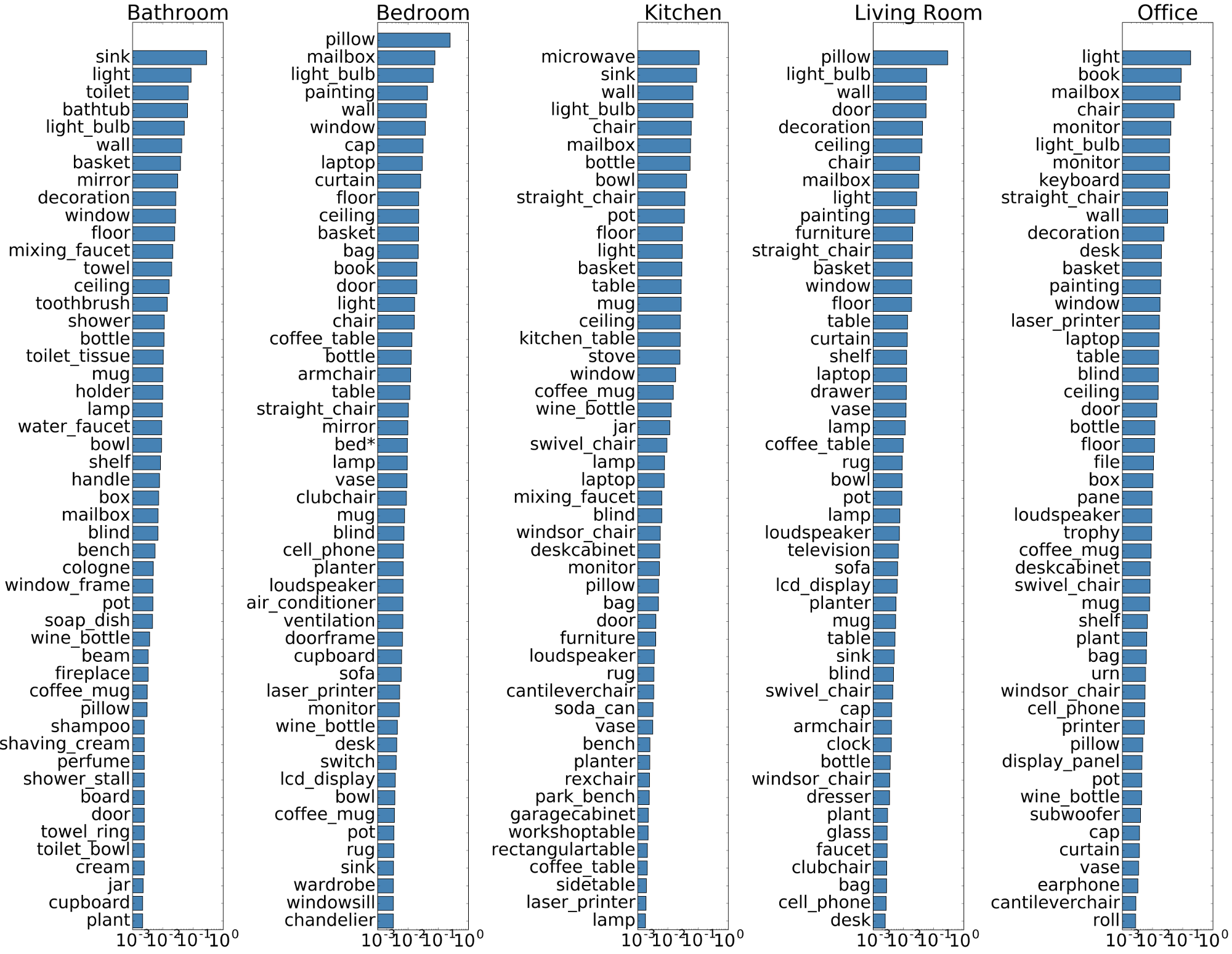}
}
\caption{Top 50 objects and their log proportions by scene type.   The unfortunate number of mailboxes is a result of a mistaken mapping of the `box' class in SUN RGB-D to a class defined as box in ShapeNets, but which contains primarily mailboxes. This is an unfortunate mishap that serves to highlight some of the difficulties inherent in working with large quantities of objects and labels in an automated way.
*beds are subdivided into a number of similar classes such as miscbeds, kingsized beds, and here we combine these into a coherent group.}
\label{fig:object_distributions_by_scene}
\end{figure*}

To create scenes, we first of all randomly choose the density of objects per square meter. In our case we have two of these densities.  For large objects we choose a density between \SI{0.1} and  \SI{0.5}{objects\per\meter\squared}, and for small objects ($<$0.4m) we choose a density between \SI{0.5} and  \SI{3.0}{objects\per\meter\squared}. Given the floor area of a scene, we then can easily calculate the number objects needed.  We sample objects for a given scene according to the distribution of objects categories in that scene-type in the SUN-RGBD real-world dataset.  We do this with the aim of including relevant objects within a context \textit{e.g.} a bathroom is more likely to contain a sink or toilet than a microwave (see Figure~\ref{fig:object_distributions_by_scene} for an object breakdown by scene type). We then randomly choose an object class according to the scene type and pick a random instance uniformly from the available models for that object type.

The objects are provided with a constant mass (10kg) and convex collision hull and positioned uniformly within the 3D space of the layouts axis aligned bounding box. To slightly bias objects towards maintaining a correctly orientated upwards direction, we offset the center of gravity on the objects to be below the mesh. Without this, we found that very few objects such as chairs were in their normal upright position after the physics simulation had completed. One drawback of the convex collision hull is that, for example, a whole table can sometimes be propped up by a small object underneath the middle of it.  

The physics engine models the movement of objects using Newtonian laws, and their interactions with each other and the layout (which is properly modelled as a static non-convex collision object). We simulate 60s of the system, leaving the objects to settle to a physically realistic configuration.  It is important to note that the scene is not necessarily organised and structured in a human manner. It contains objects in random poses and locations but the overall configuration is physically plausible \textit{i.e.} we will not have configurations where an object cannot physically support another, and unrealistic object intersections are avoided.

\section{Generating random trajectories} 
\label{ref:random trajectories}

As we aim to render videos at a large scale, it is imperative that the trajectory generation be automated to avoid costly manual labour. The majority of previous works have used a SLAM system operated by a human to collect hand-held motion: the trajectory of the camera poses returned by the SLAM system is then inserted in a synthetic scene and the corresponding data is rendered at discrete or interpolated poses of the trajectory \cite{Handa:etal:ECCV2012,Handa:etal:ICRA2014}. However, such reliance on humans to collect trajectories quickly limits the potential scale of the dataset. 

\begin{figure}[!htb]
\includegraphics[width=1.0\linewidth]{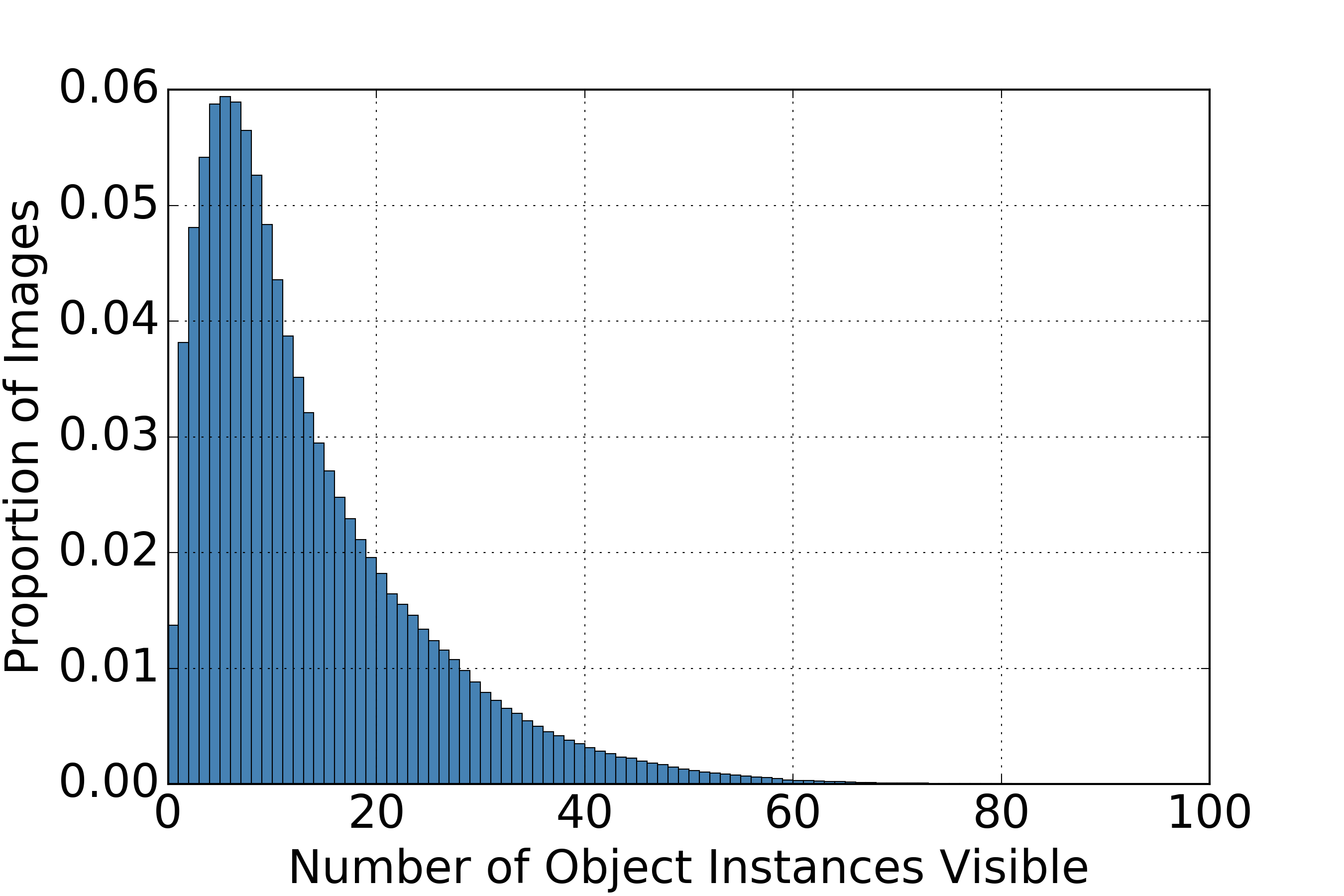}
\caption{The frequency of images with a certain number of object instances visible in the dataset.}
\label{fig:frequencyinstancevisible}
\end{figure}

We automate this process using a simple random camera trajectory generation procedure which we have not found in any previous synthetic dataset work. For our trajectories, we have the following desiderata. Our generated trajectories should be random, but slightly biased towards looking into central areas of interest, rather than, for example panning along a wall (See Figure~\ref{fig:frequencyinstancevisible} for an analysis on the number of instances visible for any given image in our final dataset). It should contain a mix of fast and slow rotations such as those of a human operator focussing on nearby and far away points.  It should also have limited rotational freedom that emphasises yaw and pitch rather than rolling, which is a less prominant motion in human trajectories.

\subsection{Two body camera trajectories}

To achieve the desired trajectory paths we simulate two physical bodies in space. One defines the location of the camera, and another, the point in space that it is focussing on as a proxy for a human focussing on random points in a scene. We take the simple approach of locking roll entirely, by setting the up vector to alway be along the positive y-axis, these two points then completely define the camera coordinate system. 

The physical approach has a number of benefits. Firstly, it provides an intuitive set of metric physical properties we can set to achieve a desired trajectory, such as the strength of the force in Newtons and the drag coefficients.  Secondly, it naturally produces smooth trajectories.  Finally, although not provided in this dataset, it automatically provides a set of IMU style accelerometer measurements, which could in future prove useful for Visual-Inertial systems.

We initialise the pose and ``look-at'' point from a uniform random distribution within the bounding box of the scene, ensuring they are less than 50cm apart. As not all scenes are convex, it is possible to initialise the starting points outside of a layout, for example in an `L'-shaped room.  Therefore, we have two simple checks. The first is to restart the simulation if either body leaves the bounding volume. The second is that within the first 500 poses at least 10 different object instances must have been visible.  This prevents trajectories external to the scene layout with only the outer wall visible.

We use simple Euler integration to simulate the motion of the bodies and apply random force vectors and drag to them independently. The body is initialized with a position, $\boldsymbol{p}$, sampled as described above, and a velocity, $\mathbf{v} = \mathbf{0}$. We begin by sampling from a uniform spherical distribution.  We achieve this by sampling from a 3-dimensional multi-variate gaussian, with $\boldsymbol{\mu} = \mathbf{0}$ and $\boldsymbol{\Sigma} = \mathbf{I}$,
\begin{equation}
\mathbf{u} \sim \mathcal{N} (\boldsymbol{\mu},\boldsymbol{\Sigma}),
\end{equation}
 we normalise $\boldsymbol{u}$ to be on the unit sphere and then scale it by a force constant, $f$, which we set to \SI{2.5}{\newton}, to arrive at our force vector $\mathbf{f}$
\begin{equation}
\mathbf{f} = f\frac{\mathbf{u}}{||\mathbf{u}||}.
\end{equation}

We also apply a drag force to dampen fast motions. We roughly model this as air drag at \SI{20}{\degree} with a \SI{30}{\cm} ball. With a cross-sectional area $A=$ \SI{0.09}{\meter \squared}, drag coefficient $C_D=0.1$, and air density $\rho=$\SI{1.204}{\kilogram \per\meter \cubed},

\begin{equation}
\mathbf{d} = -\frac{\mathbf{v}}{2||\mathbf{v}||}\rho A C_D ||\mathbf{v}||^2
\end{equation}

To calculate the acceleration we assign the body a mass, $m$, of \SI{1.0}{\kilogram}. We use simple Euler integration over a timestep, $\tau$, which here we set to $\frac{7}{300}$\SI{}{\second} for the period between shutter close and shutter open, and $\frac{1}{60}$\SI{}{\second} for the shutter open and shutter close exposure time.  

\begin{equation}
\mathbf{v}_{t} = \mathbf{v}_{t-1} + \tau\left(\frac{\mathbf{d} + \mathbf{f}}{m}\right)
\end{equation}
We also limit the maximum speed of the body to $s_{\text{max}}$, 

\[
    \mathbf{v}_{t}= 
\begin{cases}
    s_{\text{max}}\frac{\mathbf{v}_{t}}{||\mathbf{v}_{t}||},& \text{if } ||\mathbf{v}_{t}|| > s_{\text{max}}\\
    \mathbf{v}_{t},              & \text{otherwise}
\end{cases}
\]

Finally, to avoid collisions with the scene or objects we render a depth image using the z-buffer of OpenGL. If a collision occurs, the velocity is simply negated in a `bounce', which simplifies the collision by assuming the surface normal is always the inverse of the velocity vector. 

\subsection{Scene and Trajectory Description}
Listing~\ref{lst:page} shows a sample trajectory file defining the scene and camera trajectory. We provide the WordNet id, and also save the object height in meters and the full 3$\times$4 transformation of the object. 

\begin{minipage}{\linewidth}
\begin{lstlisting}[label=lst:page,language=bash, caption=Partial scene layout document after trajectory generation]
layout_file:./bedroom/bedroom3_layout.obj
object
03938244/218f86362028a45b78f8b40f4a2ae98a
wnid
03938244
scale
0.416493
transformation
0.999238  -0.00604157 -0.0385491   1.46934
0.00627241 0.999963    0.00587011 -0.0346129
0.0385122 -0.00610744  0.999239   -1.00603

object
03938244/ac2477b9b5d3e6e6c7c8ce3bef5c2aa9
wnid
03938244
scale
0.169709
transformation
 0.505633   0.123627  0.853845  3.57641
-0.00155019 0.989809 -0.142395 -0.0223919
-0.862747   0.070676  0.500672 -0.377113
....

# Poses come in alternating pairs. With shutter open
# on the first line then shutter close on the next.
# Each line has a timestamp in seconds as well as
# the camera position and look at position both defined
# in world coordinates. The layout is as follows:
# time cam_x cam_y cam_z lookat_x lookat_y lookat_z

# frame rate (Hz):      25
# shutter duration (s): 0.0166667

0.0000 -2.157 1.234 2.384  -0.5645 2.491 0.5848
0.0167 -2.157 1.233 2.384  -0.5646 2.490 0.5847

0.0400 -2.156 1.232 2.384  -0.5647 2.489 0.5843
0.0567 -2.156 1.232 2.384  -0.5648 2.489 0.5841
\end{lstlisting}
\end{minipage}

\section{Rendering photorealistic RGB frames}

The rendering engine used was version of the Opposite Renderer\footnote{\url{http://apartridge.github.io/OppositeRenderer/}} \cite{MasterThesisOppositeRenderer}, a flexible open-source ray-tracer built on top of the NVIDIA OptiX framework. We added certain extra features such as phong specular materials, ground truth materials, and multiple photon maps which can be stored in CPU memory and swapped unto the GPU. Although there were other open-source alternatives that we considered \textit{e.g.} POVRay, Blender and OpenGL, each one had their own limitations. For instance, though POVRay is able to use multi-threading on the CPU, it does not have GPU support. It is not easy to render high quality visual artefacts such as global illumination, caustics, and reflections and transparency in OpenGL and we did not find Blender as flexible for customised rendering as OptiX. 

We do not have strict real-time constraints to produce photorealistic rendering, but the scale and quality of images required does mean the computational cost is an important factor to consider. Since OptiX allows rendering on the GPU it is able to fully utilise the parallelisation offered by modern day graphics cards. This framework also provides us with significant flexibility with our rendering pipeline, enabling us to obtain ground truth information of various kinds such as depth and object instance number quite conveniently. Moreover, in future it could also allow for more complicated BRDF surface properties to be easily modelled.

\subsection{Photon Mapping}
We use a process known as photon mapping to approximate the rendering equation. Our static scene assumption makes photon mapping particularly efficient as we can produce photon maps for a scene which are maintained throughout the trajectory. A good tutorial on photon mapping is given by its creator Jensen~\textit{et al.}\cite{Jensen:etal:Course2000}.

\begin{figure}[!htb]
\centerline{
\hfill { \includegraphics[width=0.3\linewidth]{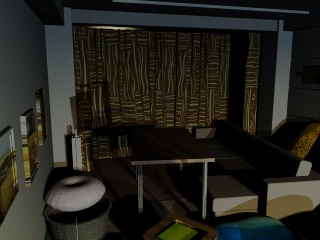} }
\hfill { \includegraphics[width=0.3\linewidth]{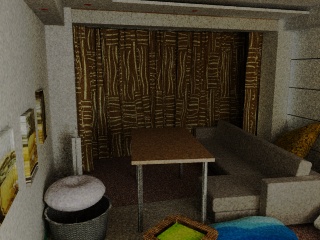} }
\hfill { \includegraphics[width=0.3\linewidth]{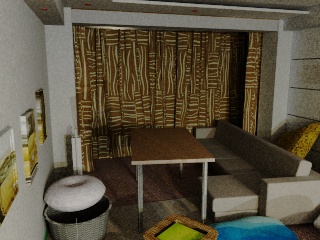} }
}
\centerline{
\hfill \makebox[0.30\linewidth][c] {\sf \footnotesize (a) Direct \& specular}
\hfill \makebox[0.30\linewidth][c] {\sf \footnotesize (b) Surface radiance}
\hfill \makebox[0.30\linewidth][c] {\sf \footnotesize (c) Combined}
}
\caption{Comparison of direct and indirect photon mappings}
\label{fig:photon_mapping}
\end{figure}

\begin{figure}[!htb]
\centerline{
\hfill { \includegraphics[width=0.49\linewidth]{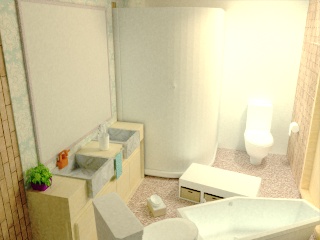} }
\hfill { \includegraphics[width=0.49\linewidth]{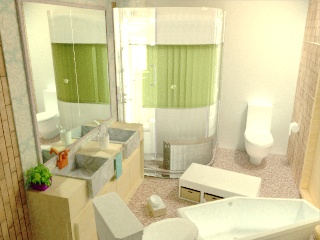} }
}
\centerline{
\hfill \makebox[0.50\linewidth][c] {\sf \footnotesize (a) No reflections \& transparency}
\hfill \makebox[0.50\linewidth][c] {\sf \footnotesize (b) With reflections \& transparency}
}
\caption{Reflections and transparency}
\label{fig:reflections}
\end{figure}

As a quick summary, this technique works via a two-pass process. In the first pass, simulated photons are emitted from light sources accumulating global illumination information and storing this information in a photon map.  In the second pass radiance information from this photon map is gathered along with direct illumination from light sources and specular reflections using ray-tracing to produce the final render, these separate and combined images can be seen in Figure~\ref{fig:photon_mapping}. Normal ray-tracing allows for accurate reflections and transparency renderings such as those in Figure~\ref{fig:reflections}, but photon mapping provides a global illumination model that also approximates indirect illumination, colour-bleeding from diffuse surfaces, and caustics (this effect can be seen through the transparent shower enclosure).

\begin{figure*}[!htb]
\centerline{
\includegraphics[width=1.\linewidth]{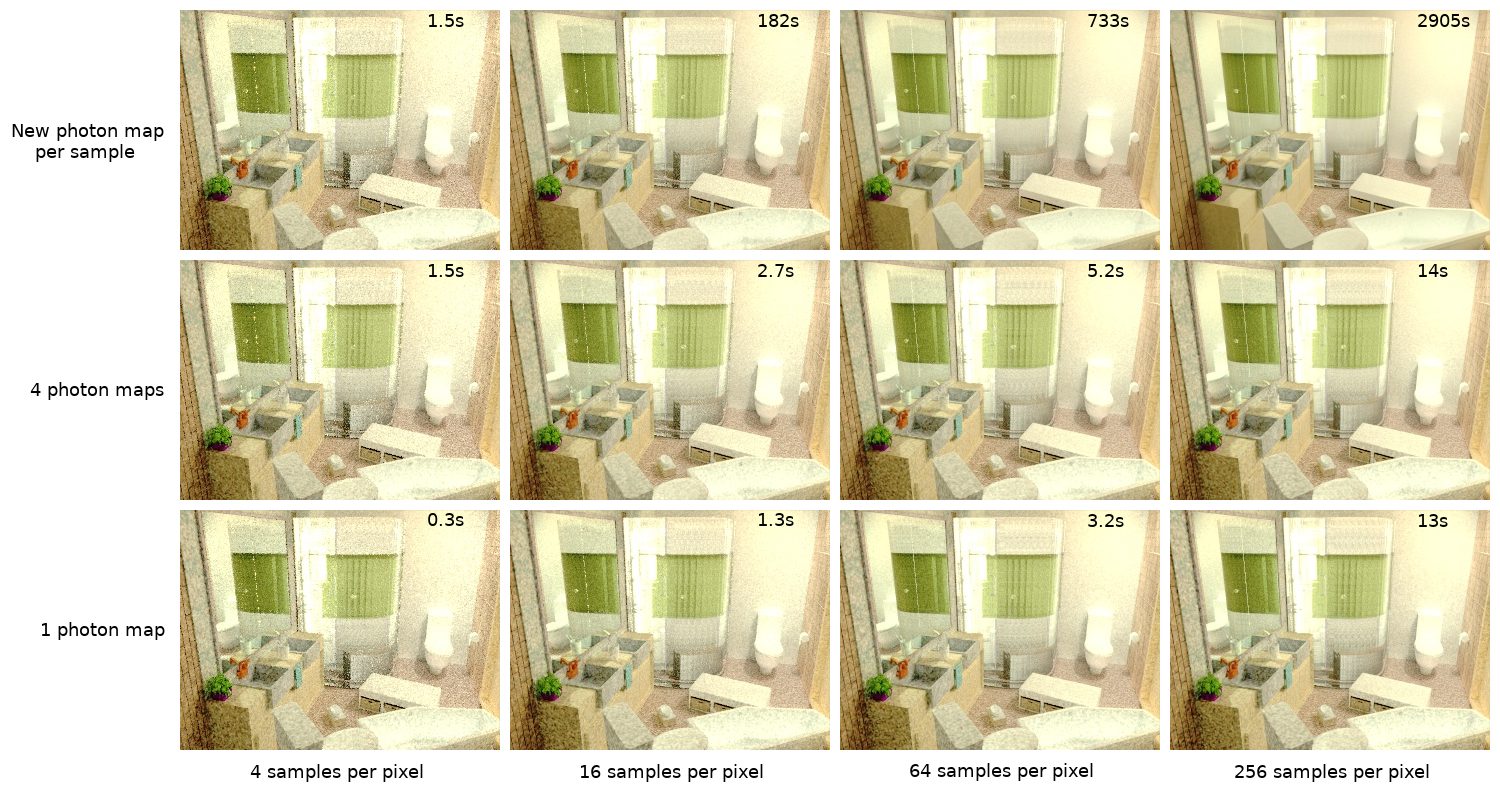}
}
\caption{Trade off between rendering time and quality. Each photon map contains approximately 3M stored photons.}
\label{fig:renderingtimequality}
\end{figure*}

\subsection{Rendering Quality} 
\label{sec:rendering_quality}

Rendering over 5M images requires a significant amount of computation.  We rendered our image on 4-12 GPUs for approximately one month.  An important tradeoff in this calculation is between the quality of the renders and the quantity of images.  Figure~\ref{fig:renderingtimequality} shows two of the most important variables dictating this balance within our rendering framework.  Our final dataset was rendered with 16 samples per pixel and 4 photon maps.  This equates to approximately 3s per image on a single GPU. 

An important threshold for the purposes of photon-mapping is that anymore than 8 photon maps exceeds the available 32GB memory. For less than 8 photon maps, we can precalculate the photon map once, and the computational cost is amortised across a trajectory. More than this and we must either store to disk or recompute a new set of photon maps for each frame in a trajectory.

\subsection{Random Layout Textures and Lighting} 

To improve the variability within our 57 layouts, we randomly assign textures to each of its constituent components.  Each layout object has a material type, which then gives a number of random texture images for that type.  For example, we have a large number of different seamless wall textures, floor textures, and curtain textures.

\begin{figure}[!htb]
\centerline{
\hfill { \includegraphics[width=0.48\linewidth]{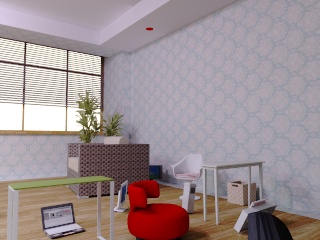} }
\hfill { \includegraphics[width=0.48\linewidth]{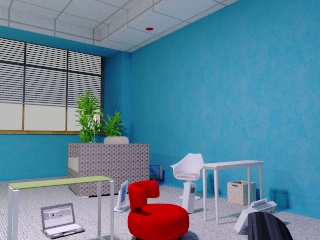} }
}
\centerline{
\hfill \makebox[0.50\linewidth][c] {\sf \footnotesize (a) Version 1}
\hfill \makebox[0.50\linewidth][c] {\sf \footnotesize (b) Version 2}
}
\caption{Different renderings of the same geometric scene with different lighting and layout textures.}
\label{fig:same_scene_diff_settings}
\end{figure}

As well as this, we add random lighting to the scene.  A number of lights between 1 and 5 is selected.  We have two types of lights, spherical orbs, which serve as point light sources and parallelograms which act as area lights. We randomly pick a hue and power of each light and then add them to a random location within the scene.  We bias this location to be within the upper half of the scene.

This approach allows an identical geometric layout to result in renders with different visual characteristics, see Figure~\ref{fig:same_scene_diff_settings}.  In this work we have only rendered a single version of each layout, however the availability of such pairs could prove an interesting facet of such randomisation in future.

\subsection{Camera Model and CRF}
\begin{figure}[!htb]
\includegraphics[width=1.0\linewidth]{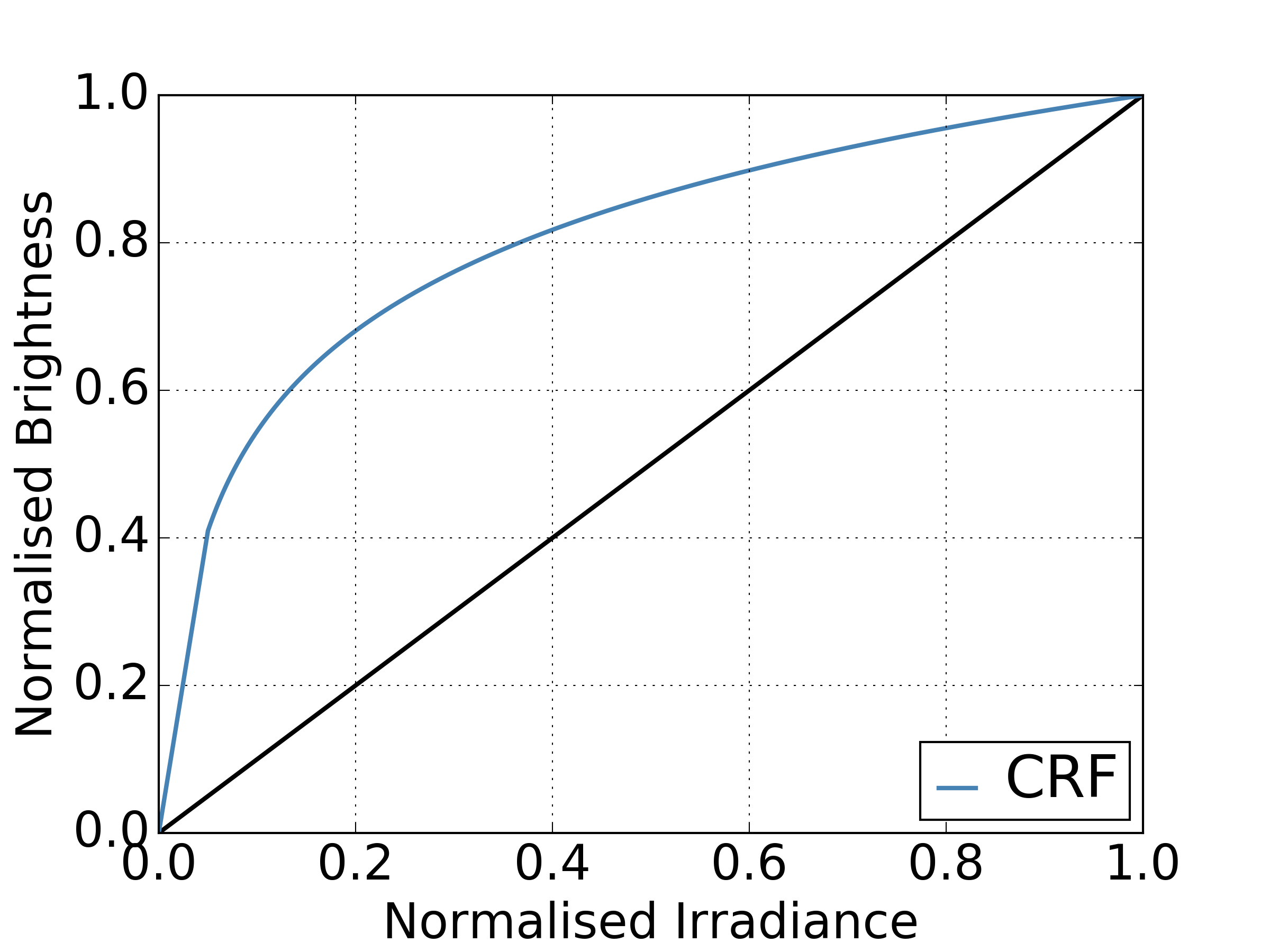}
\caption{The Camera Response Function used by our renderer.}
\label{fig:crf_plot}
\end{figure}

Our camera is a simple global shutter pinhole model, with a focal length of 20cm, a horizontal FoV of 60\degree and vertical FoV of 40\degree.  In order to make sure the rendered images are a faithful approximation to the real-world images, we also apply a non-linear Camera Response Function (CRF) that maps the irradiance to quantised brightness as in a real camera. We use a hard coded CRF in our case as shown in Figure \ref{fig:crf_plot}, however it would be relatively simple to also randomise these parameters.

\subsection{Motion Blur}
\begin{figure}[!htb]
\centerline{
\hfill { \includegraphics[width=0.5\linewidth]{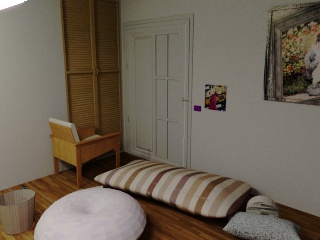} }
\hfill { \includegraphics[width=0.5\linewidth]{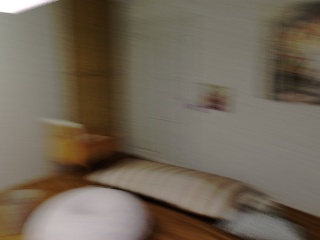} }
}
\centerline{
\hfill \makebox[0.50\linewidth][c] {\sf \footnotesize (a) without motion blur}
\hfill \makebox[0.50\linewidth][c] {\sf \footnotesize (b) with motion blur}
}
\caption{Motion blur examples.}
\label{fig:motionblur}
\end{figure}

For fast motion we integrate incoming rays throughout a shutter exposure to approximate motion blur --- this can be efficiently performed within the rendering process by changing the poses from which samples are drawn for each pixel and integrating the irradiance value rather than for example averaging RGB values after rendering. To calculate the motion blur we draw linearly interpolated lines between the camera position and look-at position at both shutter open and shutter close. Then when rendering we uniformly sample a value from $\mathcal{U}$(0,1) for different camera and ``look-at'' positions and then render in those sampled poses. For an example rendering using this technique see Figure~\ref{fig:motionblur}.

The motion blur does not affect the ground truth outputs of depth or instance segmentations.  For these images we set the pose to be the exact midpoint of the shutter exposure.

\section{Conclusion}

We have tackled the problem of producing realistic synthetic per-pixel labelled data, and we anticipate that the scale and quality of this dataset could help bridge the gap between simulations and reality and be suitable for domain adaption tasks \cite{Bousmalis:etal:NIPS2016}.  We highlight some of the problems we have successfully tackled such as physically realistic scene layouts, sensible random camera trajectory generation, and photorealistic rendering.  We also note certain areas where more work is needed.  The primary challenges still to be faced include curating accurately metric scaled objects, and insuring accurate and consistent labels on object datasets. As mentioned in Figure~\ref{fig:object_distributions_by_scene}, our automated systems mistakenly found mailboxes from ShapeNets when searching for the object category `box'. This unfortunately led to large numbers of mailboxes in indoor scenes. At present even a synthetic dataset requires significant manual intervention in cases such as this to prevent mistakes.

Although immediately useful for many computer vision tasks, the present work has a number of limitations.  Firstly, the scenes are static. This allows us to take advantage of efficient rendering techniques, but dynamic scenes, including soft bodies, would provide a more faithful representation of the real world. Secondly, we do not have certain intrinsic physical attributes of objects, such as mass or friction coefficients.  Both of these limitations mean that the dataset is not immediately applicable to active agents in an interactive physically realistic dynamic scene. However, given enough compute power, our rendering pipeline could potentially provide rendering data on-the-fly for these sorts of systems.  

The randomness inherent in our pipeline also allows for a continuous stream of unseen training examples, dynamically designed to target current limitations of a model being trained. In the future, it is likely that the generation of training data and the training of models will become more tightly interleaved, and the advantages of automatically generated training data becomes clear.

\section{Acknowledgements}

Research presented in this paper has been supported by Dyson Technology Ltd. We would also like to thank Patrick Bardow for providing optical flow code.

{\small
\bibliographystyle{ieee}
\bibliography{egbib}
}
\end{document}